\documentclass[a4paper]{spie}  %>>> use for US letter paper
\usepackage{amsmath,amsfonts,amssymb}
\usepackage{graphicx,subfigure}
\usepackage[colorlinks=true, allcolors=blue]{hyperref}
\DeclareMathOperator*{\argmin}{arg\,min}

\title{Low-Dose CT via Deep CNN with Skip Connection and Network in Network}

\author[a]{Chenyu You}
\author[a]{Linfeng Yang}
\author[b]{Yi Zhang}
\author[c]{Ge Wang}
\affil[a]{Department of Bioengineering, Stanford University, USA 94305}
\affil[b]{College of Computer Science, Sichuan University, China 610065}
\affil[c]{Department of Biomedical Engineering, Rensselaer Polytechnic Institute, USA 12180}

\begin{document} 
\maketitle

\begin{abstract}
A major challenge in computed tomography (CT) is how to minimize patient radiation exposure without compromising image quality and diagnostic performance. The use of deep convolutional (Conv) neural networks for noise reduction in Low-Dose CT (LDCT) images has recently shown a great potential in this important application. In this paper, we present a highly efficient and effective neural network model for LDCT image noise reduction. Specifically, to capture local anatomical features we integrate Deep Convolutional Neural Networks (CNNs) and Skip connection layers for feature extraction. Also, we introduce parallelized $1\times 1$ CNN, called Network in Network, to lower the dimensionality of the output from the previous layer, achieving faster computational speed at less feature loss. To optimize the performance of the network, we adopt a Wasserstein generative adversarial network (WGAN) framework. Quantitative and qualitative comparisons demonstrate that our proposed network model can produce images with lower noise and more structural details than state-of-the-art noise-reduction methods.
\end{abstract}


\section{INTRODUCTION}
\label{sec:intro}  % \label{} allows reference to this section

X-ray computed tomography is widely used for clinical screening, diagnosis, and intervention. However, the radiation dosage associated with CT examinations may potentially induce some cancerous and genetic diseases~\cite{de2009projected}. As a result, the well-known ALARA~\cite{brenner2007computed} (as low as reasonably achievable) principle is universally accepted in practice, reducing unnecessary radiation exposure during medical CT imaging. One of the commonly-used methods is to lower the X-ray flux towards the x-ray detector array by adjusting the milliampere-seconds (mAs) and kVp settings for data acquisition. However, since CT imaging is a quantum integration process, an insufficient amount of photons will introduce excessive statistical noise and significantly deteriorate image quality. Therefore, how to preserve image quality for clinical tasks at minimum radiation dose has been one of the major endeavors in the CT field over the past decade.

Deep learning (DL) has been now applied in almost all medical tomographic imaging areas, inspired by a large amount of image processing results~\cite{wang2016perspective,wang2017machine}. In particular, several DL-based studies for image noise reduction were performed~\cite{wolterink2017generative,yang2017low,kang2016deep,you2018ct,lyu2018super}. Since CNN models learn  high-level representations in terms of multiple layers of feature abstraction from big training images, it is expected to have a better denoising capability than other classic image-domain methods. In this paper, we aim  to maintain anatomical and pathological information, and at the same time suppress image noise due to low radiation dose. Specially, we develop a new ConvNet architecture for LDCT denoising. In order to progressively capture both local and global anatomical features, here we design cascaded subnetworks to integrate complementary textural information. Moreover, by introducing residual learning at the image reconstruction stage, the network model is made to learn the residuals between a bicubic interpolation image and the corresponding full-dose CT (FDCT) image so that  the denoising performance can be boosted. Finally, with parallelized CNNs (Network in Network) local patches within the receptive field are effectively analyzed~\cite{lin2013network}. As far as the loss function is concerned, we introduce the $L_{1}$ norm instead of $L_{2}$ distance to disencourage blurring~\cite{you2018low}.

\begin{figure*}[htb]

\begin{minipage}[b]{1.0\linewidth}
  \centering
  \centerline{\includegraphics[width=0.8\linewidth]{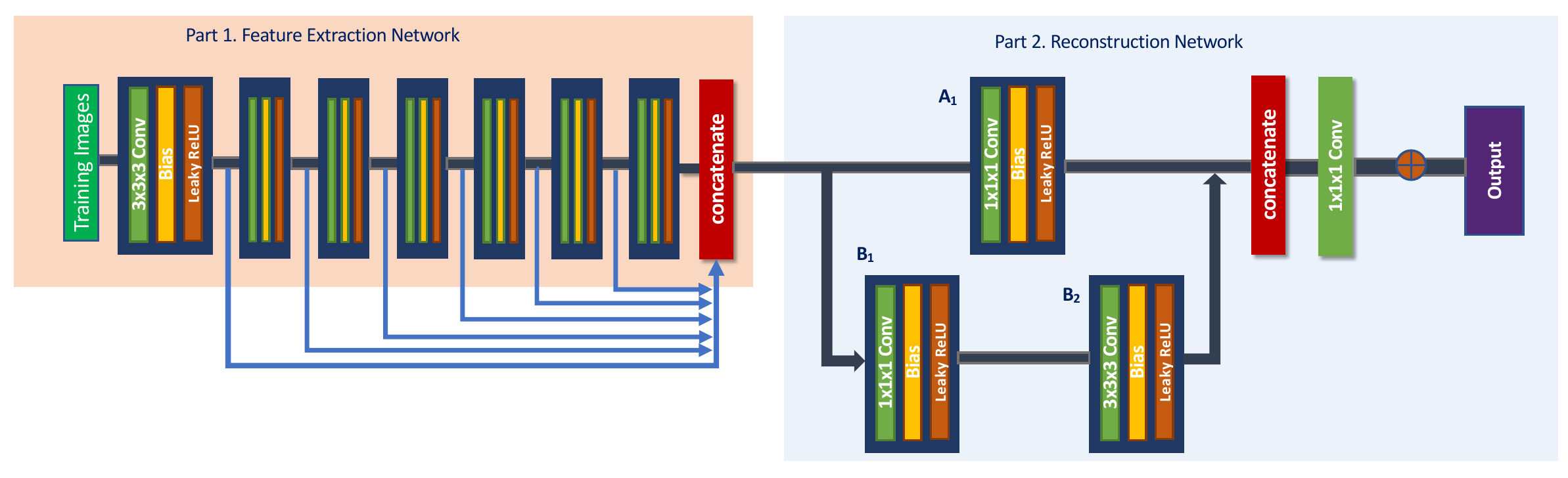}}
%  \vspace{2.0cm}
  \centerline{(a)Architecture of the generator $G$}\medskip
\end{minipage}
\begin{minipage}[b]{1.0\linewidth}
  \centering
  \centerline{\includegraphics[width=0.8\linewidth]{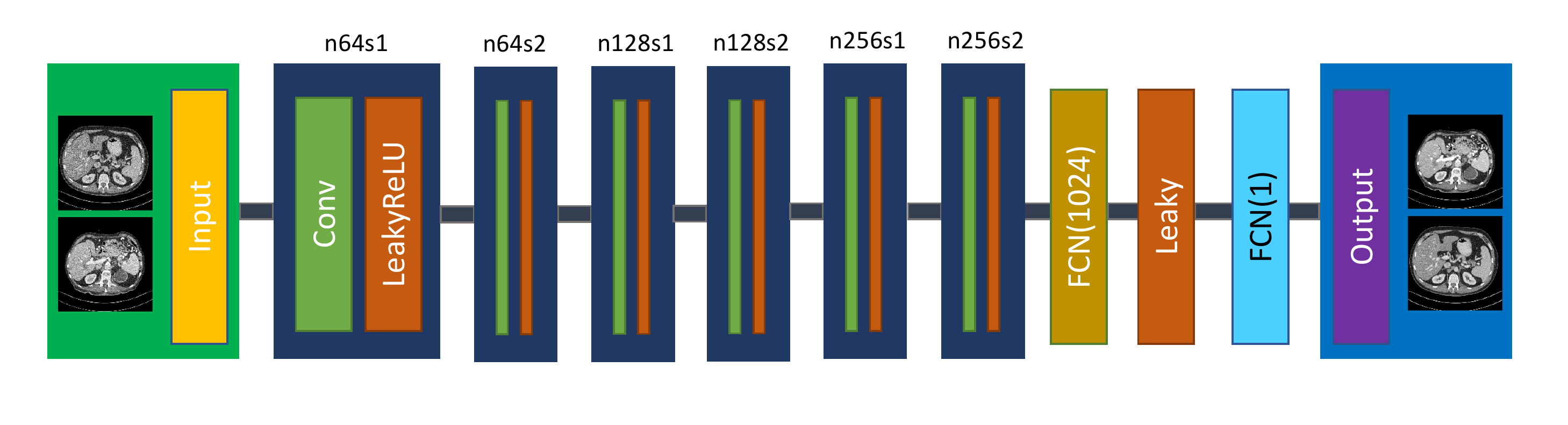}}
%  \vspace{2.0cm}
  \centerline{(b)Architecture of the discriminator $D$}\medskip
\end{minipage}
\caption{Our proposed network structure. The $G$ is composed of a feature extraction network and a reconstruction network. Note that the reconstruction block $A_{1}$ is denoted as Channel $A$, the reconstruction blocks $B_{1}$ and $B_{2}$ as Channel $B$, $n$ stands for the number of convolutional kernels, and $s$ for convolutional stride. For example, $n32s1$ means that the convolutional layer has 32 kernels with stride 1.}
\label{fig:all}
\end{figure*}

\section{Methods}

Let a vector $\pmb{x} \in \mathbb{R}^{M \times 1}$ represent a noisy LDCT image of $N \times N$ pixels, and a vector $\pmb{y} \in \mathbb{R}^{M \times 1}$ its corresponding NDCT image, $M = N \times N$. A DL-based network model $\boldsymbol{DL}$ with the multiple processing layers is trained to process LDCT images  according to a non-linear input-output mapping, which is equivalent to solve the following optimization problem:
\begin{equation}
\argmin_{\boldsymbol{DL}}||\boldsymbol{DL}(\pmb{x}) - \pmb{y}||_{1}
\end{equation}

Our network constitutes two components: the generative model $G$ and the discriminative model $D$ as shown in Fig.~\ref{fig:all}. In the feature extraction network, the number of filters are $32, 26, 22, 18, 14, 11, 8$ for the Conv layers respectively. Also, in the image reconstruction network, the three channels are cascaded. The reconstruction block $A_{1}$, $B_{1}$, $B_{2}$ consist of $24, 8, 8$ filters respectively. Because all the outputs from the feature extraction layers were densely connected, and the final outputs after reconstruction is large, therefore we introduce $1\times 1$ CNN after the reconstruction network to reduce the input dimension and decrease computational complexity. Instead of constructing a high-quality image by the network itself, we incorporate residual learning strategy to capture high-frequency features that can help improve the quality of low-dose CT images~\cite{yu2017computed}.

\begin{figure*}[!ht]
\centering
%\begin{subfigure}{0.24\textwidth}
\subfigure[Full Dose]{\includegraphics[width=0.30\textwidth]{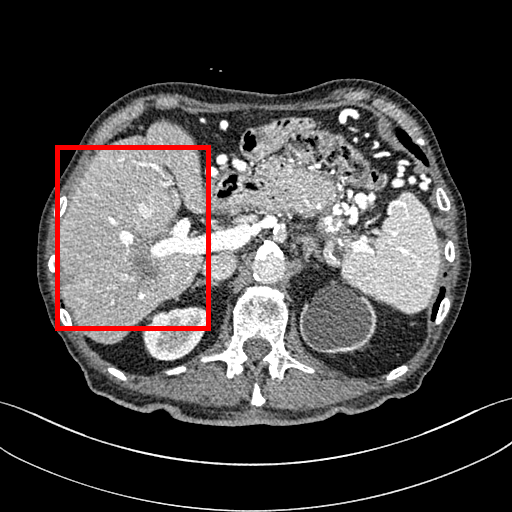}\label{fig: full1}}
\subfigure[Quarter Dose]{\includegraphics[width=0.30\textwidth]{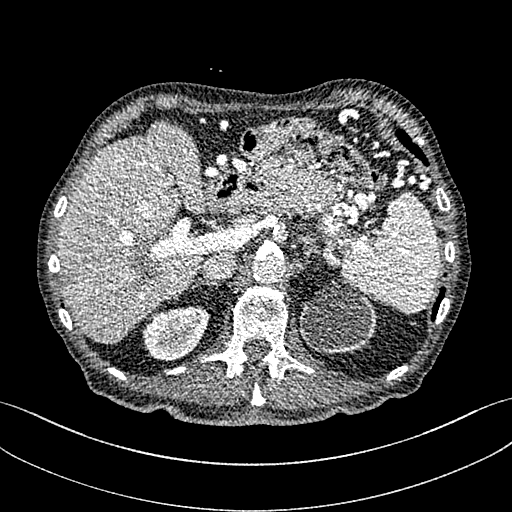}\label{fig: quarter1}}
\subfigure[CNN-L1]{\includegraphics[width=0.30\textwidth]{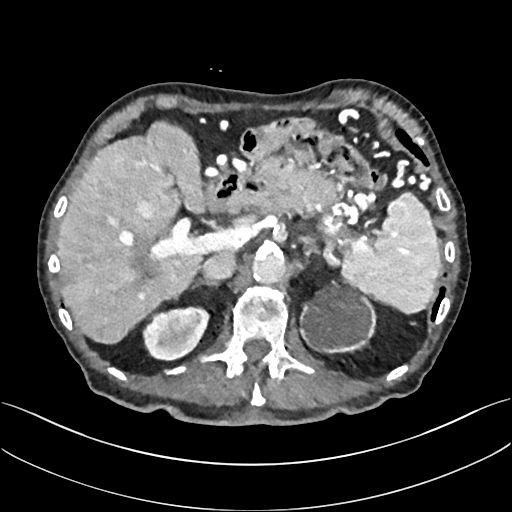}\label{fig: rmse1}}

\subfigure[WGAN]{\includegraphics[width=0.30\textwidth]{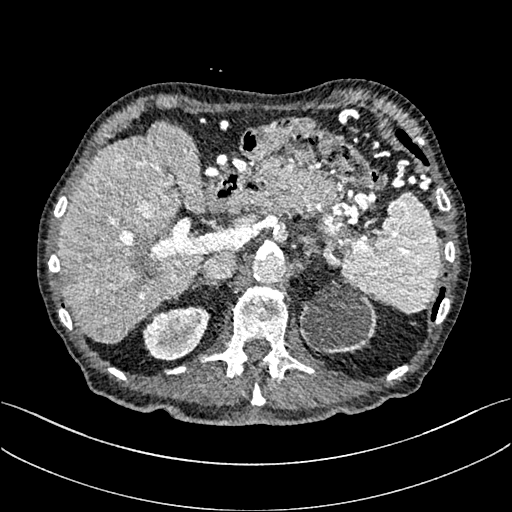}\label{fig: wgan1}}
\subfigure[DCSCN]{\includegraphics[width=0.30\textwidth]{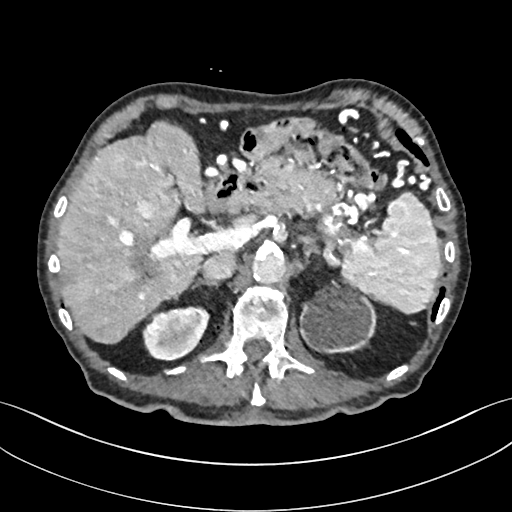}\label{fig: dcscn1}}
\subfigure[DCSWGAN]{\includegraphics[width=0.30\textwidth]{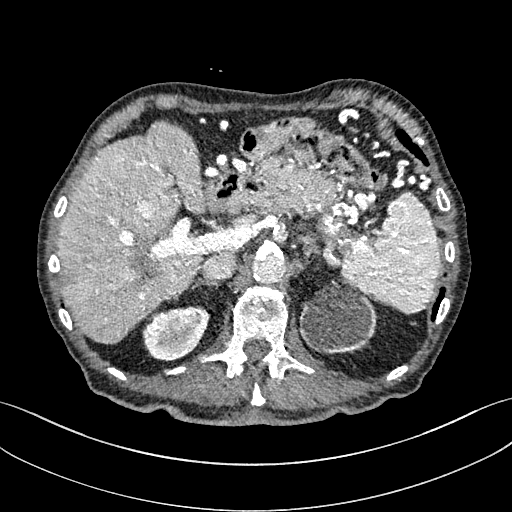}\label{fig: dcwgan1}}
\caption{Results with abdomen CT images.(a) FDCT, (b) LDCT, (c) CNN-L1, (d) WGAN, (e) DCSCN, and (h) DCSWGAN. The red box indicates the region zoomed in Fig.~\ref{fig: example1_roi}. This display window is [-160, 240]HU.}
\label{fig: example1}
\end{figure*}

\begin{figure}[!ht]
\centering
%\begin{subfigure}{0.24\textwidth}
\subfigure[Full Dose]{\includegraphics[width=0.15\textwidth]{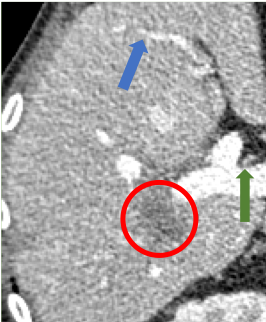}\label{fig: full1_roi}}
\subfigure[Quarter Dose]{\includegraphics[width=0.15\textwidth]{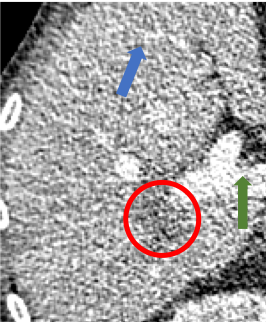}\label{fig: quarter1_roi}}
\subfigure[CNN-L1]{\includegraphics[width=0.15\textwidth]{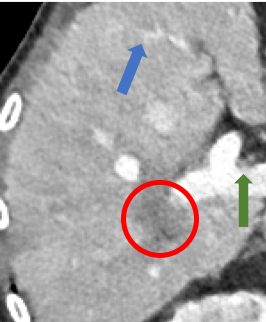}\label{fig: rmse1_roi}}
\subfigure[WGAN]{\includegraphics[width=0.15\textwidth]{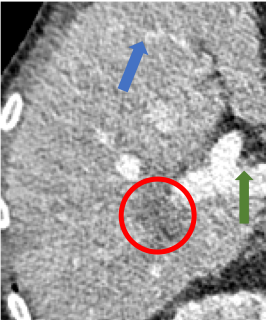}\label{fig: wgan1_roi}}
\subfigure[DCSCN]{\includegraphics[width=0.15\textwidth]{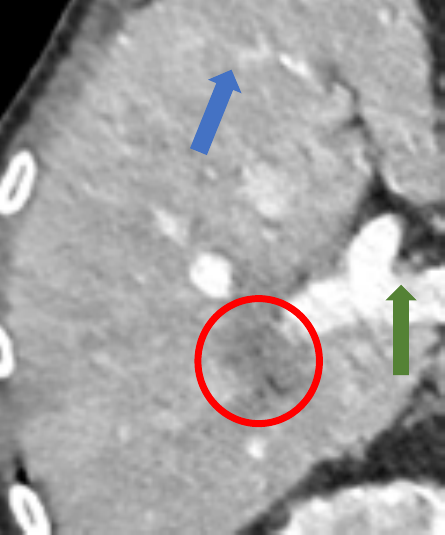}\label{fig: dcscn_roi1}}
\subfigure[DCSWGAN]{\includegraphics[width=0.15\textwidth]{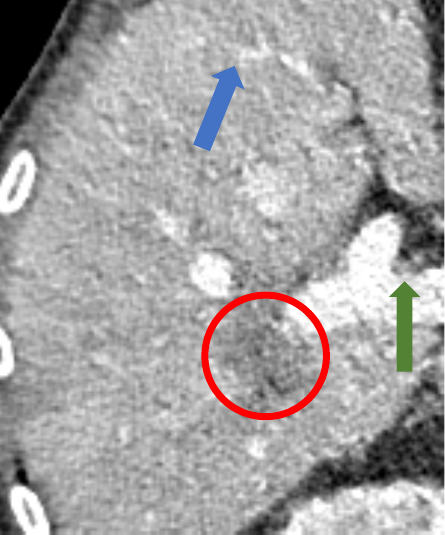}\label{fig: dcwgan_roi1}}

\caption{Zoomed  regions of interest(ROIs) marked by the red box in Fig.~\ref{fig: example1}. (a) FDCT, (b) LDCT, (c) CNN-L1, (d) WGAN, (e) DCSCN, and (f) DCSWGAN. The dashed circle shows the metastasis, and the green and blue arrows indicate two subtle structural features. The display window is [-160,240]HU}
\label{fig: example1_roi}
\end{figure}

% \begin{table}[!ht]
% \renewcommand{\arraystretch}{1.3}
% \centering
% \caption{Number of filters on each convolution (conv) layer of the generative network.}
% \begin{tabular}{c c c c c c c c c c c c} \\
% \hline\hline
% & \multicolumn{7}{c}{Feature extraction network} & \multicolumn{4}{c}{Reconstruction network} \\
% & 1 & 2 &3 & 4 &5 &6 &7 &A1 &B1 &B2 &Output \\
% \cline{2-12}
% $G$ & 32 & 26 & 22 & 18 & 14 & 11 & 8 & 24 & 8 & 8 & 1\\
% \hline\hline
% \end{tabular}
% \label{table: generator_num_filter}
% \end{table}

The covariance of pixel level features will significantly influence the denoising performance~\cite{you2018low}. Indeed, in our experiments the pure CNN-based model tends to produce blurry features. GANs~\textit{et al.}~\cite{goodfellow2014generative} is a promising approach to address the aforementioned limitations, since GAN is a framework for generative modeling of data through minimizing the discrepancy between the prior data distribution $P_z$ of the generated outputs from $G$ and the real data distribution $P_r$. Hence, we force the denoised image to stay on the image manifold by matching the distribution of real images to that of synthesized input images. Even though GAN has been widely applied in image processing, they suffer from model divergence and are unstable to train~\cite{radford2015unsupervised}. To regularize the training process for GAN, we adopt the Earth Moving
distance (EM distance), instead of the original Jensen-Shannon (JS) divergence, in the objective function~\cite{arjovsky2017wasserstein}. Thus, the adversarial loss is formulated as:
\begin{equation}
    \min_{G}\max_{D}L_{\mathrm{WGAN}}(D,G) = -\mathbb{E}_{\pmb{y}}[D(\pmb{y})]+\mathbb{E}_{\pmb{x}}[D(G(\pmb{x}))]+\lambda \mathbb{E}_{\hat{\pmb{x}}}[(||\nabla_{\hat{\pmb{y}}}D(\hat{\pmb{y}})||_2-1)^2],
\label{eq: loss_DG_WGAN}
\end{equation}
where the first two terms are for the Wasserstein estimation, the third term penalizes the deviation of the gradient
norm with respect to the input from one, $\hat{\pmb{y}}$ is uniformly sampled along straight lines pairs of denoised and real images, and $\lambda$ is a regularization parameter.

\begin{figure*}[!t]
\centering
%\begin{subfigure}{0.24\textwidth}
\subfigure[Full Dose]{\includegraphics[width=0.30\textwidth]{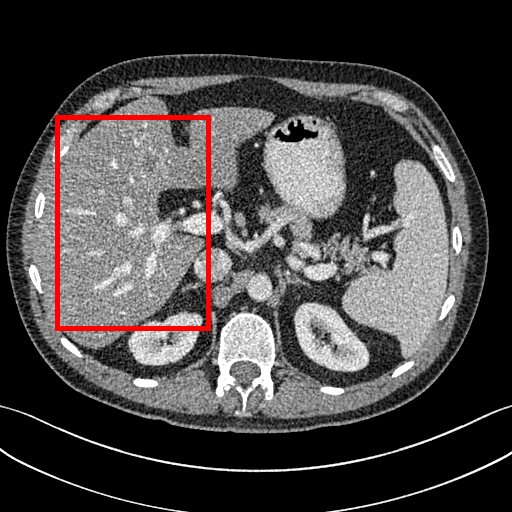}\label{fig: full2}}
\subfigure[Quarter Dose]{\includegraphics[width=0.30\textwidth]{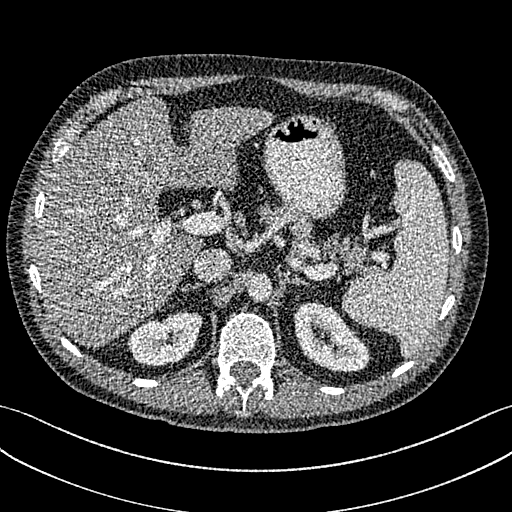}\label{fig: quarter2}}
\subfigure[CNN-L1]{\includegraphics[width=0.30\textwidth]{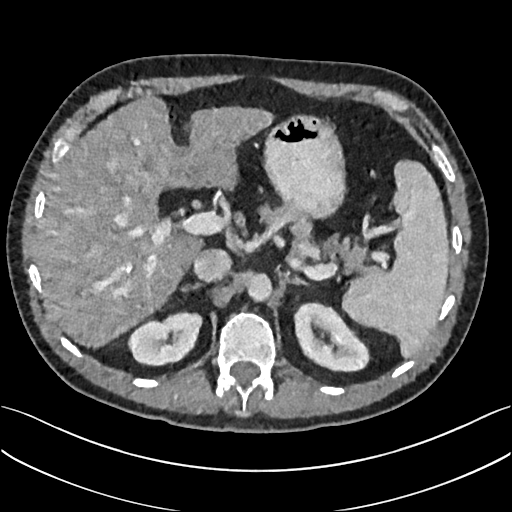}\label{fig: rmse2}}
\subfigure[WGAN]{\includegraphics[width=0.30\textwidth]{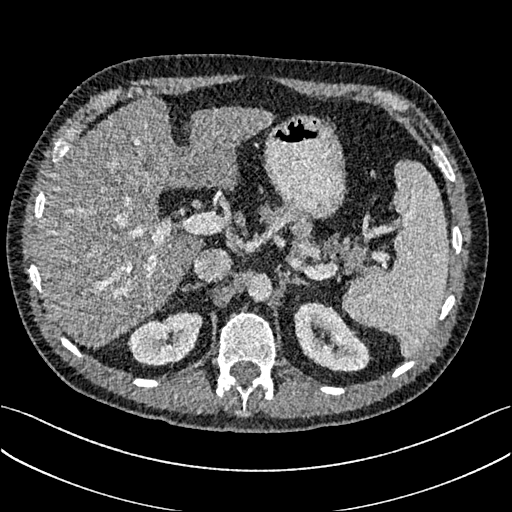}\label{fig: wgan2}}
\subfigure[DCSCN]{\includegraphics[width=0.30\textwidth]{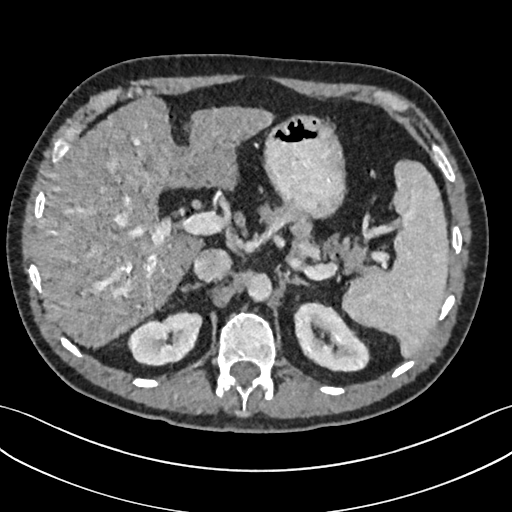}\label{fig: dcscn2}}
\subfigure[DCSWGAN]{\includegraphics[width=0.30\textwidth]{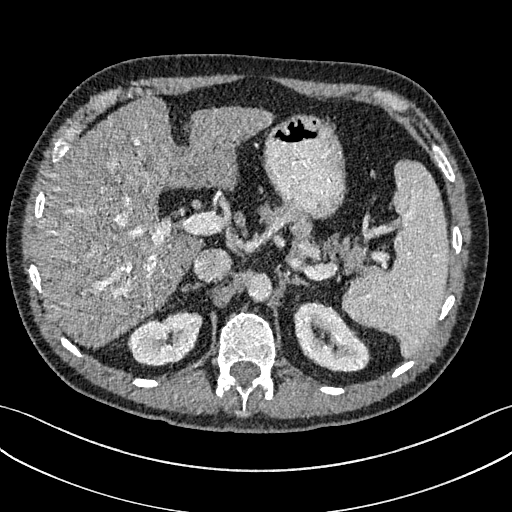}\label{fig: dcwgan2}}

\caption{Results from abdomen CT images.(a) FDCT, (b) LDCT, (c) CNN-L1, (d) WGAN, (e) DCSCM, and (f) DCSWGAN. The red box indicates the region of interest zoomed in Fig.~\ref{fig: example3_roi}. This display window is [-160, 240]HU.}
\label{fig: example3}
\end{figure*}

\begin{figure}[!t]
\centering
%\begin{subfigure}{0.24\textwidth}
\subfigure[Full Dose]{\includegraphics[width=0.15\textwidth]{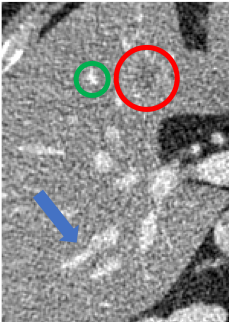}\label{fig: full2_roi}}
\subfigure[Quarter Dose]{\includegraphics[width=0.15\textwidth]{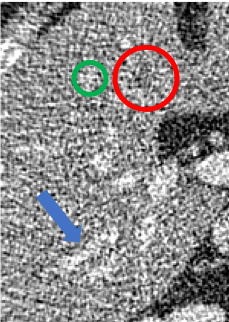}\label{fig: quarter2_roi}}
\subfigure[CNN-L1]{\includegraphics[width=0.15\textwidth]{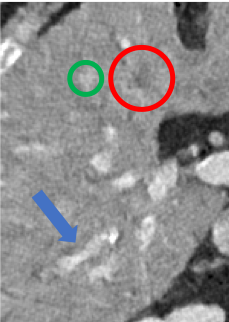}\label{fig: rmse2_roi}}
\subfigure[WGAN]{\includegraphics[width=0.15\textwidth]{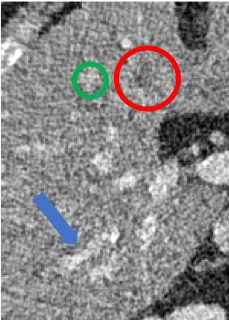}\label{fig: wgan2_roi}}
\subfigure[DCSCN]{\includegraphics[width=0.15\textwidth]{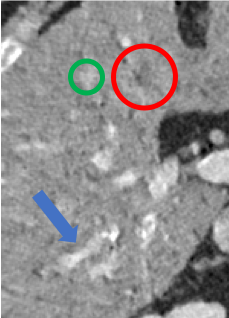}\label{fig: dcscn_roi2}}
\subfigure[DCSWGAN]{\includegraphics[width=0.15\textwidth]{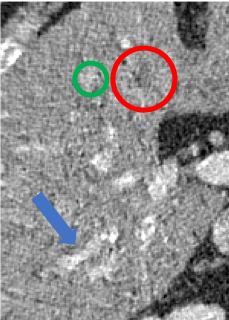}\label{fig: dcwgan_roi2}}
\caption{Zoomed regions of interest (ROIs) marked by the red box in Fig.~\ref{fig: example3}. (a) FDCT, (b) LDCT, (c) CNN-L1, (d) WGAN, (e) DCSCM, and (f) DCSWGAN. The dashed circle indicates the metastasis, and the green and blue arrows show two subtle  structural features. The display window is [-160,240]HU}
\label{fig: example3_roi}
\end{figure}

Although $L_{1}$ and $L_{2}$ losses are both the mean-based loss function, the effects of these two loss functions differ in terms of denoising. Compared with the $L_{2}$ loss, the $L_{1}$ loss neither over-penalize large differences nor tolerate small errors between denoised images and the gold-standard. Thus, the $L_{1}$ loss alleviates some limitations of the $L_{2}$ loss. Additionally, the $L_{1}$ loss shares the same merits that the $L_{2}$ loss has; e.g, a fast convergence speed.

The $L_{1}$ loss is formulated as follows:
\begin{equation}
L_{\mathrm{1}}(G) = \frac{1}{HWD}\mid G(\pmb{x}) -\pmb{y}\mid
\label{eq: L{1}_norm}
\end{equation} 
where $H$, $W$, $D$ stand for the height, width, and depth of a 3D image patch, respectively, $\pmb{y}$ denotes a gold-standard image (NDCT), and $G(\pmb{x})$ represents a denoised image from a LDCT image $\pmb{x}$.

Besides, there are two aspects in the sparse representation
step for image denoising, which are the prior information level and the sparsity level. We first introduce the adversarial loss to capture local anatomical information. Then, we use $L_{1}$ loss to improve the sparsity of our representation, leading to the solution of the following optimization problem.

Leveraging Eqs.~(\ref{eq: loss_DG_WGAN}) and~(\ref{eq: L{1}_norm}) together, the overall joint objective function is formulated as:
\begin{equation}
L_{\mathrm{obj}} = \min_{G}\max_{D} L_{\mathrm{WGAN}}(D,G)+\lambda_1 L_{\mathrm{1}}(G)
\label{eq: loss_overall}
\end{equation}
where $\lambda_1$ is a regularization parameter to balance the information preservation and the sparsity-promotion between the WGAN adversarial loss and the $L_{\mathrm{1}}$ loss.

%------------------------------------------------------------------------
\section{EXPERIMENTAL RESULTS}
To evaluate the effectiveness of the proposed method, we compared it with existing state-of-the-art denoising methods, including CNN-L1~($L_{1}$-net)~\cite{you2018low} and WGAN-based CNN~\cite{yang2017low}. Note that all the parameters of these selected benchmark methods were set to that suggested in the original papers. For brevity, we denote our Deep CNN with Skip Connection and Network in Network as DCSCN, and the model using a Wasserstein Generative Adversarial Network as DCSWGAN.

The experiment set-up is as follows. First, to minimize the generalization error, we adopted leave-one-out cross-validation to refine the denoising performance. Then, in the training phase, $499,996$ pairs of image patches of size $80\times 80$ from 7 patients were randomly selected. For validation, $5,096$ pairs of image patches were extracted from other 3 patients and set to the same size. It is worth noting that the size of extracted patches was made large enough to include regions of liver lesions. Next, in addition to preserve the integrity of data, here we scaled the CT Hounsfield Value (HU) to the unit interval [0,1] before the images were fed to the network. Finally, we used three common image quality metrics: peak signal-to-noise ratio (PSNR), structural similarity index (SSIM)~\cite{wang2003multiscale}, and root-mean-square error (RMSE) to evaluate the denoised image quality.

\begin{table}[t]
\renewcommand{\arraystretch}{1.1}
\renewcommand\tabcolsep{2pt}
\centering
\caption{Quantitative results associated with different approaches in Figs.~\ref{fig: example1} and~\ref{fig: example3}.}
\begin{tabular}{|c|c|c|c|c|c|c|c|}
\hline
& \multicolumn{3}{c|}{Fig.~\ref{fig: example1}} && \multicolumn{3}{c|}{Fig.~\ref{fig: example3}}
\\
\cline{2-4}\cline{6-8}
& PSNR & SSIM & RMSE && PSNR & SSIM & RMSE \\
\cline{2-4}\cline{6-8}
LDCT & 22.818 & 0.761 & 0.0723 && 21.558 & 0.659 & 0.0836 \\
CNN-L1 & 27.791 & 0.822 & 0.0408 && 26.794 & 0.738 & \textbf{0.0457} \\
WGAN & 25.727 & 0.801 & 0.0517 && 24.655 & 0.711 & 0.0585 \\
DCSCN & \textbf{28.016} & \textbf{0.883} & \textbf{0.0397} && \textbf{26.943} & 0.730 & 0.0530 \\
DCSWGAN & 26.928 & 0.828 & 0.0449 && 25.721 & 0.808 & 0.0517 \\
\hline
\end{tabular}
\label{table: psnr&ssim&rmse}
\end{table}

The visual inspection of our results indicates that the LDCT images in Figs.~\ref{fig: quarter1} and~\ref{fig: quarter2} have strong background noises. Furthermore, we find that the $l_1$-net has a great noise suppression capability, but it still has over-smoothing effects on some textural details in the ROIs in Fig.~\ref{fig: rmse1_roi}. The $l_1$-net achieved a high signal-to-noise ratio (SNR), but it yielded lower contrast resolution. From ROIs in Fig.~\ref{fig: rmse2_roi}, it is seen that there are still some blocky effects marked by the blue arrow. Figs.~\ref{fig: wgan1} and~\ref{fig: wgan2} display the WGAN-processed denoised LDCT images with improving structural identification. However, as shown in Figs.~\ref{fig: wgan1_roi} and~\ref{fig: wgan2_roi}, the WGAN model also introduced strong image noise. In Figs.~\ref{fig: dcscn1} and~\ref{fig: dcscn2}, the proposed DCSCN achieved noise reduction but also suffered from image blurring . As shown in Fig.~\ref{fig: dcwgan1} and~\ref{fig: dcwgan2}, our proposed DCSWGAN network model demonstrates the best performance in noise reduction and feature preservation as compared to all the competing denoisng methods. Figs.~\ref{fig: dcwgan_roi1} and~\ref{fig: dcwgan_roi2} illustrate that DCSWGAN not only effectively suppressed strong noise but also kept subtle textural information, outperforming other denoising models;  see ROIs (in Figs.~\ref{fig: example1_roi} and~\ref{fig: example3_roi}) and/or zoom in for better visualization.

The PSNRs, SSIMs, and RMSEs are listed in Table~\ref{table: psnr&ssim&rmse}. For noise reduction, the performance metrics were significantly improved by our proposed method (DCSCN). This demonstrates that using residual learning  steak artifacts and image noise can be largely removed, enhancing the image quality. In this pilot study, DCSCN achieved the best performance in terms of PSNR and SSIM, and preserved anatomical features the most faithfully. However, there still exits blurry effects as shown in Figs.~\ref{fig: example1_roi} and ~\ref{fig: example3_roi}. DCSWGAN obtained the second best results in term of SSIM. It is noted that our method DCSWGAN produced visually pleasant results with sharp edges.

\section{Conclusion}
In this work, we have proposed a CNN-based network with skip-connection and network in network to capture structural information and suppress image noise. First, both local and global features are cascaded through skip connections before passing to the reconstruction network. Then, multi-channels are introduced for the reconstruction network with different local receptive fields to optimize the reconstruction performance. Also, the network in network technique is applied to lower the computational complexity. Our results have suggested that the proposed method could be generalized to various medical image denoising problems but further efforts are needed for training, validation, testing, and optimization.

% References
\bibliography{report} % bibliography data in report.bib
\bibliographystyle{spiebib} % makes bibtex use spiebib.bst

\end{document}